\documentclass[11pt]{article}
\usepackage[utf8]{inputenc}

\usepackage[preprint]{acl}
\usepackage{subcaption}
\usepackage{booktabs}
\usepackage{amsmath}
\usepackage{tablefootnote}

\usepackage{times}
\usepackage{latexsym}

\usepackage[T1]{fontenc}
\usepackage{listings}
\lstdefinelanguage{json}{
  basicstyle=\ttfamily\small,
  numbers=none,
  breaklines=true,
  breakatwhitespace=true,
  showstringspaces=false,
  frame=single,
  rulecolor=\color{black!30},
  tabsize=2,
  columns=fullflexible,
  keepspaces=true,
  literate=
   *{0}{{{\color{black}0}}}{1}
    {1}{{{\color{black}1}}}{1}
    {2}{{{\color{black}2}}}{1}
    {3}{{{\color{black}3}}}{1}
    {4}{{{\color{black}4}}}{1}
    {5}{{{\color{black}5}}}{1}
    {6}{{{\color{black}6}}}{1}
    {7}{{{\color{black}7}}}{1}
    {8}{{{\color{black}8}}}{1}
    {9}{{{\color{black}9}}}{1}
}

\usepackage[utf8]{inputenc}

\usepackage{microtype}

\usepackage{inconsolata}

\usepackage{graphicx}
\usepackage{enumitem}

\usepackage[breakable]{tcolorbox}
\usepackage[dvipsnames]{xcolor}

\tcbuselibrary{breakable,skins}

\title{SOP-Maze: Evaluating Large Language Models on Complicated Business Standard Operating Procedures}

\author{
    Jiaming Wang$^{1,*}$, 
    Zhe Tang$^{1,*}$, 
    Zehao Jin$^{2,*}$, 
    Hefei Chen$^{1,*}$, 
    Yilin Jin$^{1,*}$, 
    Peng Ding$^{3,*}$, \\
    Xiaoyu Li$^{1}$, 
    Xuezhi Cao$^{1}$
    \thanks{$^*$ Equal contribution.}
    \thanks{$^1$ Meituan M17. \texttt{\{ wangjiaming15, tangzhe03, chenhefei02, jinyilin, lixiaoyu28, caoxuezhi \}@meituan.com}}
    \thanks{$^2$ Georgia Institute of Technology. \texttt{zjin350@gatech.edu}}
    \thanks{$^3$ Nanjing University. \texttt{dingpeng@smail.nju.edu.cn}}
}

\begin{document}
\setlist[itemize]{nosep, leftmargin=*}

\maketitle
\begin{abstract}
As large language models (LLMs) are widely deployed as domain-specific agents, many benchmarks have been proposed to evaluate their ability to follow instructions and make decisions in real-world scenarios.
However, business scenarios often involve complex standard operating procedures (SOPs), and the evaluation of LLM capabilities in such contexts has not been fully explored. To bridge this gap, we propose SOP-Maze, a benchmark constructed from real-world business data and adapted into a collection of 397 instances and 3422 subtasks from 23 complex SOP scenarios. We further categorize SOP tasks into two broad classes: Lateral Root System (LRS), representing wide-option tasks that demand precise selection; and Heart Root System (HRS), which emphasizes deep logical reasoning with complex branches. Extensive experiments reveal that nearly all state-of-the-art models struggle with SOP-Maze. We conduct a comprehensive analysis and identify three key error categories: (i) route blindness: difficulty following procedures; (ii) conversational fragility: inability to handle real dialogue nuances; and (iii) calculation errors: mistakes in time or arithmetic reasoning under complex contexts. The systematic study explores LLM performance across SOP tasks that challenge both breadth and depth, offering new insights for improving model capabilities. We have open-sourced our work on: \url{https://github.com/meituan-longcat/SOP-Maze}.
\end{abstract}

\section{Introduction}

In recent years, LLMs have made significant progress in natural language understanding and generation~\citep{kumar2024large,qin2024large,wang2025history}. The milestone is the significant improvement in instruction-following ability~\citep{wei2021finetuned,ouyang2022training,wang2023selfinstruct,zhang2025recommendation, li2025otter}, which has driven their gradual deployment as professional agents in domain-specific applications.

\begin{figure}[t]
  \includegraphics[width=\columnwidth]{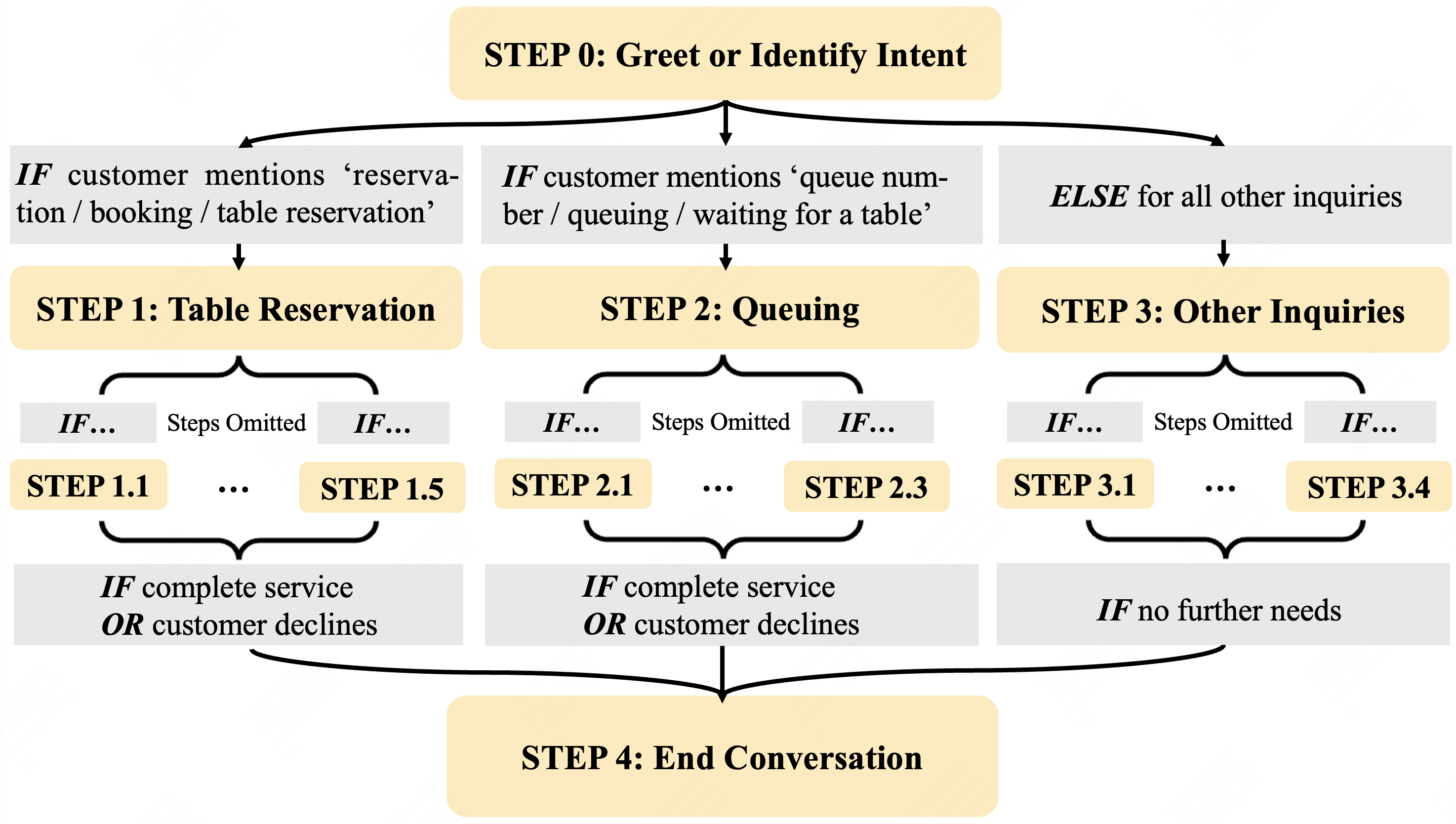}
  \caption{An example of business SOPs. }
  \label{fig:SOP} 
\end{figure}

To evaluate this capability, considerable achievements have been made in generic and compositional instruction scenarios~\citep{single_benchmark_COLLIE,multi_benchmark_Parrot,multi_benchmark_StructFlowBench}. However, many real-world applications of LLMs, such as business scenarios, are driven by more complex standard operating procedures (SOPs)~\citep{grohs2023largelanguagemodelsaccomplish,kourani2024leveraginglargelanguagemodels}. SOPs define a standard route of thinking or task execution for models involving conditional branches or complex constraints, are more challenging than regular instruction-following tasks (Figure~\ref{fig:SOP}); meanwhile, the model is expected to comply with procedures and produce the required output robustly, even when given noisy input. This reveals a significant gap between existing benchmarks and requirements from business SOPs.

\begin{figure*}[t]
    \centering
    \includegraphics[width=0.9\textwidth]{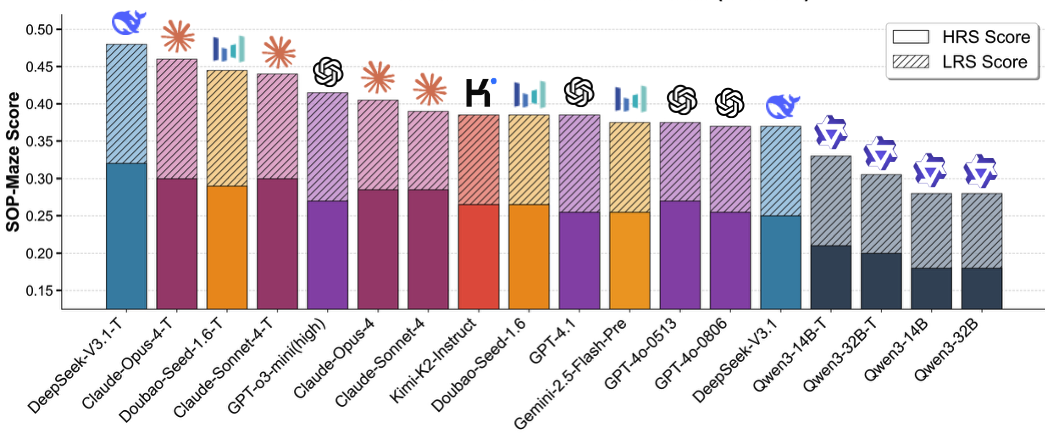}
    \caption{Leaderboard result on SOP-Maze. Each bar reports the SOP-Maze score ( average of the HRS and LRS scores ) for a model.}
    \label{fig:leaderboard}
\end{figure*}

\begin{figure*}[t]
  \includegraphics[width=\textwidth]{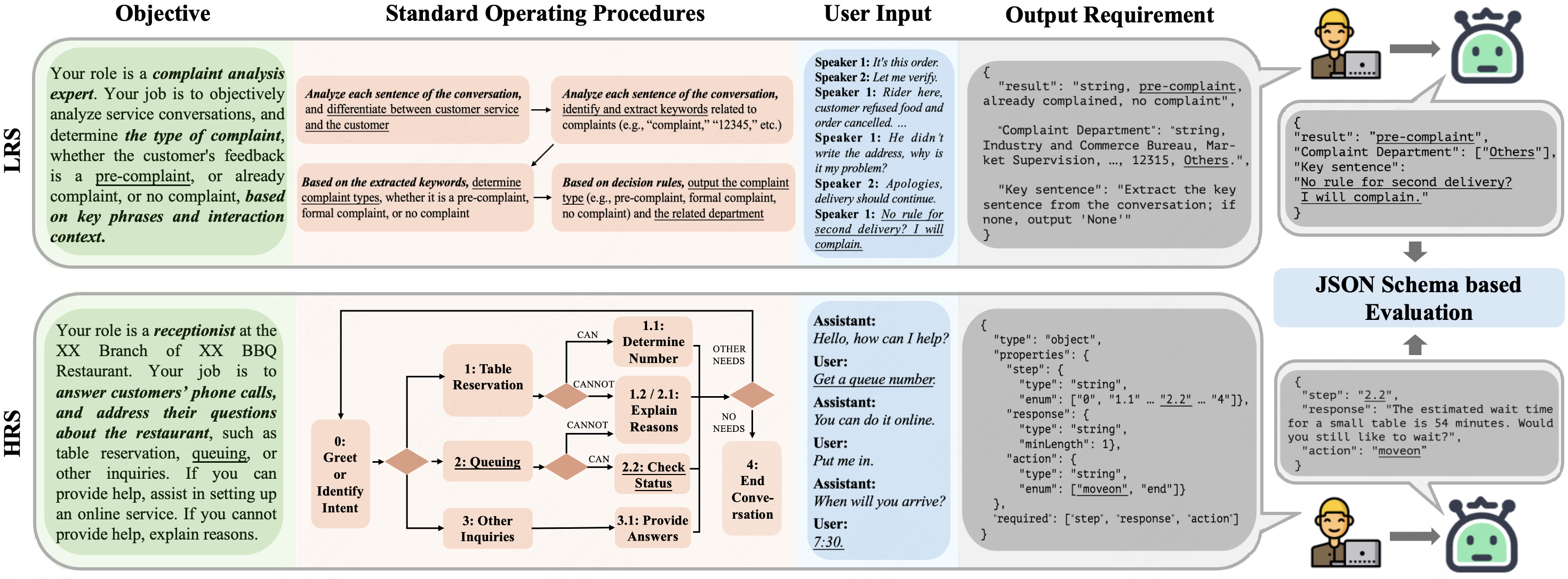}
  \caption{Illustration of SOP-Maze. Based on the context and characteristics of the SOPs, business SOP tasks are categorized into two types, LRS and HRS. Each task prompt comprises 4 key components: Objective, Standard Operating Procedures, User Input and Output Requirement. After the LLM generates an output, it is assessed using JSON Schema based Evaluation.}
  \label{fig:main}
\end{figure*}

To this end, this paper introduces SOP-Maze, a benchmark constructed from real-world business data. SOP-Maze is developed through human-in-the-loop refinement, resulting in 397 instances across 23 distinct business scenarios; each SOP task incorporates a nested or intertwined execution logic route, with an average specification length of 5,040 tokens. We innovatively organize SOP tasks in SOP-Maze into two categories, based on the context and characteristics of the SOPs. As illustrated in Figure~\ref{fig:main}, \textbf{Lateral Root System (LRS)}\footnote{\label{fn:root}\textbf{Lateral Root System} and \textbf{Heart Root System} are terms borrowed from plant root morphology in botany.} represents wide SOP tasks with relatively shallow branching structure, requiring the model to make accurate choice among alternatives with procedure compliance. \textbf{Heart Root System (HRS)}\textsuperscript{\ref{fn:root}} captures deep SOP tasks characterized by long and intricate logical chains, in this case, successful task completion requires the model to faithfully traverse a multi-step reasoning path and preserve contextual consistency until reaching the final decision. Together, LRS and HRS probe LLM capabilities across both breadth and depth dimensions of complex business SOPs under noisy conditions. To the best of our knowledge, SOP-Maze is the first benchmark evaluating the practical instruction-following capability of models in business SOP scenarios.

Based on SOP-Maze, we conduct a comprehensive evaluation of 18 prominent LLMs. To ensure reliability, we adopt a JSON schema based evaluation that is model-free, highly efficient, and capable of producing precise scores. Extensive experiments reveal that the benchmark poses substantial challenges, and prominent LLMs consistently fall short of following complex business SOPs, providing new insights into the practical capabilities of LLMs. Our contributions are as follows:

1. We propose SOP-Maze, the first SOP benchmark for real-world business scenarios, concluding 397 difficult tasks derived from 23 real-world business scenarios.

2. We categorize SOP tasks into two types: LRS and HRS, to evaluate LLM capabilities across both the breadth and depth dimensions of complex business SOP scenarios.

3. Extensive experiments on 18 SOTA models reveal the gap between current LLM capabilities and the requirements of business SOPs. Our analysis reveals three key limitations: (i) route blindness; (ii) conversational fragility; and (iii) calculation errors, providing new insights into model capabilities.

\section{Related works}
\textbf{LLM Benchmarks for Instruction Following} \quad As LLMs are increasingly used in real-world tasks, instruction-following capability has become a key metric for evaluating model practicality~\citep{single_benchmark_IFEVAL,single_benchmark:InFoBench,multi_benchmark_Multi-IF}. Early works focused on single-turn dialogues with simple constraints (e.g. semantic and format constraints)~\citep{zhou2023instruction, xia2024fofo, tang2024struc}, which limited their applicability. Later works moved toward more real-world scenarios: CELLO~\citep{he2024can} generated complex instructions from task descriptions and users' text input, Complexbench~\citep{wen2024benchmarking} synthesized instructions using real-world data, adding constraints based on fixed patterns. Furthermore, Guidebench~\citep{diao2025guidebench} raises task difficulty by incorporating domain-specific conditions and system feedback. However, these approaches lack a complex and realistic standard operation procedures in instructions, which is a critical factor in complex business tasks and pose challenges to evaluating LLMs in business scenarios.

\textbf{LLM Benchmarks for Business}  \quad Prevailing benchmarks primarily focus on Business Process Management (BPM), evaluating LLMs in tasks like modeling and optimizing complex business processes~\citep{berti2024pmllmbenchmark, fahland2024howwell, kourani2024evaluatinglargelanguagemodels}. For instance, $\mathrm{BP}^{\mathrm{C}}$~\citep{fournier2024benchmarkcausalbusinessprocess} proposes a benchmark to assess LLM capabilities in causal reasoning and explaining decision points within business processes. SAPM~\citep{10680677} evaluates LLMs on semantic-aware tasks, such as anomaly detection and next activity prediction in process mining. 

Furthermore, existing procedural benchmarks like SOPBench~\citep{li2025sopbenchevaluatinglanguageagents} and SOP-Bench~\citep{nandi2025sopbenchcomplexindustrialsops} evaluate language agents through tool-calling and API execution. SOPBench primarily assesses procedural compliance by verifying if agents execute mandatory prerequisite helper functions before a service action. Similarly, SOP-Bench emphasizes industrial workflows, testing robustness against ambiguous instructions and redundant tool sets in task-oriented automation.

In contrast, SOP-Maze shifts the challenge from external tool manipulation to deep logical traversal within SOP structures. Unlike the tool-centric nature of the aforementioned benchmarks, SOP-Maze evaluates whether LLMs can faithfully follow long logical chains, handle the nuances of noisy multi-turn conversations (e.g., sarcasm and intent reversal), and perform context-aware calculations—all while maintaining strict adherence to complex, nested execution routes.

\section{The SOP-Maze Benchmark}

\subsection{SOP-Maze Overview}
\begin{figure}[t]
  \includegraphics[width=\columnwidth]{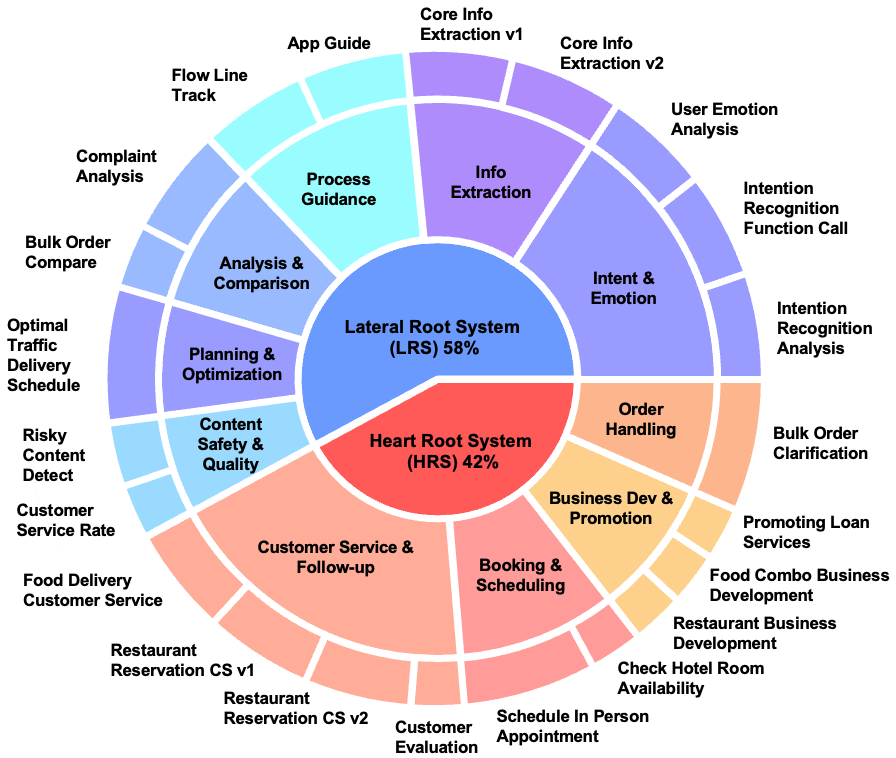}
  \caption{SOP-Maze instances distribution.}
  \label{fig:datadistribution}
\end{figure}

SOP-Maze is designed to evaluate LLM capabilities to effectively assist users in complex real-world business SOP scenarios. SOP-Maze encompasses 23 business scenarios (10 in HRS and 13 in LRS), comprising a total of 397 instances and 3422 subtasks. Comprehensive statistics are presented in Figure~\ref{fig:datadistribution}.

\subsection{Data Curation Process}
As shown in Figure~\ref{fig:main}, each task prompt in SOP-Maze comprises 4 key components:

1. \textbf{Objective} defines the background, model role, and task objectives.

2. \textbf{Standard Operating Procedures} specifies the execution logic and operation details.

3. \textbf{User Input} represents the user-provided data requiring processing.

4. \textbf{Output Requirement} mandates adherence to a predetermined output format specification.

The following subsections demonstrate the four stages involved in the construction of the SOP-Maze dataset: (i) Data collection, (ii) SOP refinement, and (iii) User input choice, followed by (iv) the statement about benchmark quality control.

\subsection{Data Collection}
With user consent, we collect 300,000 data records from our own API router logs, capturing internal business department invocations of LLM for business task execution with user consent. Following rule-based filtering to control data volume, we perform semantic clustering on the dataset. Based on clustering results, we manually select the 23 largest clusters—representing the most frequently occurring business scenarios—to constitute the 23 scenarios of SOP-Maze.

\subsection{Benchmark Quality Control}
\textbf{SOP Refinement:} In practical user scenarios, SOPs are incrementally developed through step-by-step completion. Consequently, certain instances exhibit logical chain inconsistencies and self-contradictions. We preserve the logical inconsistencies to evaluate model robustness under real-world data conditions, while resolving self-contradictions through user communication and requirement validation to ensure the uniqueness and accuracy of ideal solutions.

\textbf{User Input Collection:} We systematically analyze model performance under identical scenarios (same SOP) by validating with users in business department and confirm that the selected scenarios and user inputs authentically reflected the pain points they encounter when using LLMs in real-world applications. This process yielded a final dataset of 397 instances with 3422 subtasks.

\textbf{Dataset Quality Control:} To ensure data integrity, we recruit five professional annotators holding bachelor's degrees with over three years of annotation experience. These annotators conduct cross-validation to identify and rectify lethal logical inconsistencies in SOPs and user prompts, ensuring each task yields a single definitive answer. Each task undergoes quality inspection by a minimum of three annotators, requiring unanimous consensus before inclusion in the final dataset.

\textbf{Scoring Quality Control:} Scoring validation employs a cross-verification framework where each task is evaluated by at least three annotators who assess responses generated by the three models Claude-4-Sonnet~\citep{LLM_claude_sonnet4}, GPT-4.1~\citep{LLM_gpt4_1} and Deepseek-R1~\cite{deepseekai2025deepseekr1incentivizingreasoningcapability}. Only tasks achieving cross-validation consistency across all evaluators are incorporated into the final dataset.

\section{Task Formulation}
In this section, we present the task format of SOP-Maze and outline the evaluation criteria.
\subsection{Task Format}
SOP-Maze incorporates a comprehensive evaluation framework comprising two core components: task composition and evaluation protocol. Given that open-ended tasks such as agent-chat may require LLM-based evaluation, which introduces uncontrollable factors to our benchmark, we employ a reference-based evaluation approach to mitigate this challenge.

In authentic AI customer service scenarios, SOPs provide tested models with detailed guidance, and users expect models to generate responses directly based on these SOPs. However, such responses are inherently difficult to evaluate, as models' creative variations preclude simple string matching. Evaluation typically relies on semantic similarity computation (unreliable) or LLM-as-a-judge approaches (costly and unstable). To balance the trade-off among evaluation cost, reliability, and scenario authenticity, we assign indices to all reference outputs in SOP guidance. For tasks requiring free-form responses, we mandate that models simultaneously output both the user-expected response and the corresponding reference index from the SOP guidance. This design preserves the natural response generation that users require while enabling the benchmark to perform reliable and efficient evaluation through simple index matching.

\subsection{Evaluation Criteria}
\label{sec:evaluation_criteria}
As outlined in the preceding section, the model is required to generate responses conforming to the JSON schema specified within the prompt. The JSON schema typically encompasses various output format constraints (e.g., boolean, string types) and incorporates optional parameters for certain keys. A baseline score of 0.2 points is awarded when the model output adheres to the prescribed format requirements. Complete alignment between model output and the reference answer yields the maximum score of 1.0 points.

\begin{table}[h]
\small
\centering
\begin{tabular}{l|c|c|c}
\toprule
Total & \multicolumn{3}{c}{397} \\
\midrule
Score & 0.2 & 0 & 1 \\
\midrule\midrule
DeepSeek-V3.1-Thinking & 268 & 7 & 132 \\
Claude-Opus-4-Thinking & 227 & 39 & 131 \\
Doubao-Seed-1.6-ThinkingOn & 243 & 27 & 127 \\
\bottomrule
\end{tabular}
\caption{Performance of Top Three Models}
\label{tab:performance}
\end{table}
\vspace{-0.3cm}

\section{Experiments}
\label{sec:exp}
The results are presented in Figure~\ref{fig:leaderboard} and detailed in Table~\ref{tab:maintable1} and Table~\ref{tab:maintable2}. The experimental configuration is detailed below. We comprehensively assess 18 LLMs, including 11 API-based models and 7 open-source models~\citep{LLM_claude_sonnet4,LLM_dsv3,LLM_gpt4_1,LLM_doubao2024,LLM_gemini_flash_2024,LLM_o3mini,LLM_kimik2,LLM_qwen3}. All models are set to default hyper-parameters as demonstrated on API configuration page or huggingface.

As mentioned in Section~\ref{sec:evaluation_criteria}, models are evaluated using a three-tier scoring system:

\begin{equation*}
\small
    S = \begin{cases}
        1.0 & \text{correct response} \\
        0.2 & \text{valid format, incorrect response} \\
        0 & \text{invalid format}
    \end{cases}
\end{equation*}

This scoring framework ensures that models are evaluated on both procedural compliance (format adherence) and substantive performance (content accuracy), with greater weight assigned to correctness of responses.

\section{Analysis}
\label{sec:analysis}

\begin{table*}[t]
\small
\centering
\begin{tabular}{l|c|c|c}
\toprule
Score & \multicolumn{3}{c}{0.2} \\
\midrule
Error Type & Route Blindness & Conversational Fragility &  Calculation Error   \\
\midrule\midrule    
DeepSeek-V3.1-Thinking & 177 & 166 & 60 \\
Claude-Opus-4-Thinking & 151 & 149 & 60 \\
Doubao-Seed-1.6-ThinkingOn & 164 & 145 & 63 \\
\bottomrule    
\end{tabular}
\caption{Error breakdown for the top three models on \textbf{format-correct} responses (score=0.2). A single response may exhibit multiple error types.}
\label{tab:breakdown}
\end{table*}

To understand why models struggle on SOP-Maze beyond formatting issues, we focus on the responses that already satisfy the required JSON schema (i.e., the baseline score $0.2$; see Section~\ref{sec:evaluation_criteria}) but still fail to match the reference decision.

To ensure the objectivity and exhaustiveness of our error taxonomy, we followed an iterative qualitative procedure: 

(1) \textbf{Exploratory Case Study}: Three independent annotators analyzed all failures from the top-three models, generating free-text descriptions of error patterns without predefined categories.

(2) \textbf{Taxonomy Standardization}: These descriptions were clustered into a unified taxonomy. We established a formal annotation SOP to minimize subjectivity and ambiguity during the coding process.

(3) \textbf{Validation}: We applied this taxonomy to additional models and non-reasoning modes (see Section~\ref{sec:think_vs_no_think}). No significant new error types emerged, suggesting the taxonomy is exhaustive for current LLM failures in SOP tasks.

(4) \textbf{Reliability Check}: A separate annotator cross-verified 50 sampled cases per model. Any disagreements were resolved through consensus-based discussion to refine the final counts.

Through this process, we identified three dominant failure categories: \textbf{route blindness}, \textbf{conversational fragility}, and \textbf{calculation error}. We quantify their frequencies for the top-three models in Table~\ref{tab:breakdown} (categories are non-exclusive). In the following subsections, we analyze these categories and how they are triggered by the structural properties of SOP-Maze.




In the following subsections, we analyze these three error categories and explain how they are triggered by the structural properties of SOP-Maze.

\subsection{Fail to Grasp the Full Context of SOPs}
A central source of failure is that models do not reliably execute the SOP as a \emph{procedure}: they deviate to an incorrect branch, skip prerequisite checks, or apply a less specific rule when a more specific exception is required. We refer to this family of errors as \textbf{Route Blindness}. Importantly, it manifests differently in LRS and HRS due to their distinct graph structures.

\textbf{Route Blindness in LRS: }
LRS scenarios are shallow in depth (at most three levels from root to leaves) but wide in branching: each parent node can have more than 10 child nodes. On average, an LRS scenario contains about 5 parent nodes and 58 leaf nodes. This width creates a large set of plausible next steps that must be compared \emph{in parallel}. We observe that models often commit early to an incorrect branch and fail to recover, suggesting difficulty in maintaining and pruning a large candidate set under procedural constraints.

\textbf{Route Blindness in HRS: }
HRS scenarios are deeper and emphasize prerequisite satisfaction over long horizons. Here we observe a different pattern: models frequently \emph{skip} intermediate nodes and jump to later steps. Inspecting the reasoning traces of reasoning models suggests the skips are not random; rather, models often \emph{misjudge} that a precondition has already been met, and then proceed as if the subsequent step were valid. This behavior indicates that even reasoning models can be over-confident in implicit state tracking and lack robust self-correction when the SOP requires explicit verification.


\subsection{Fail to Grasp the Nuances of Daily Conversation}
SOP-Maze inputs are grounded in multi-turn conversations that reflect real user behavior. We find that models are brittle to three recurring conversational phenomena---\textbf{strong contextual reliance}, \textbf{disfluent phrasing}, and \textbf{subtle intent}---which we summarize as \textbf{conversational fragility}. These phenomena are not mere ``style'' issues: they directly affect which SOP branch is applicable.

\textbf{Strong Contextual Reliance: }
In realistic dialogues, user intent evolves across turns, and later utterances can override earlier ones. Models, however, often anchor on the initial instruction and underweight subsequent updates. In the ``Intention Recognition Analysis'' scenario, the user first threatens, ``\textit{If you don't do anything about this, I'm going to file a complaint,}'' but later concludes with ``\textit{I'll let it slide this time.}'' Although this is a clear reversal, models frequently fail to revise the inferred intent and continue the SOP as if the complaint were still pending.

\textbf{Disfluent Phrasing: }
Daily conversation is noisy: irregular turn-taking and interleaved signals can disrupt basic comprehension. For instance, in a driving scenario, navigation prompts are mixed with the user's speech:
``\textit{Assistant: Hello, is this Ms. Zhang? User: \underline{Hello,} make a U-turn... \underline{uh, yes}. [...Omitting middle rounds...] Assistant: So, are you interested? User: Current speed is 126, \underline{yes}, you are speeding.}''
Here, the underlined parts are the user's actual speech. Models often fail to isolate the short but decisive affirmation (``yes'') and instead treat the entire utterance as navigation instructions, which leads to an incorrect downstream step selection.

\textbf{Subtle Intent: }
Models also struggle with non-literal language (e.g., sarcasm, irony), where surface meaning contradicts speaker intent. In the ``Food Combo Business Development'' scenario, a user responds:
``\textit{Haha, yeah, right. Sounds amazing. Looks like that company is back at it again, trying to get their hands on our paychecks.}''
This is a sarcastic rejection, but models frequently interpret it as positive feedback and proceed with an inappropriate sales-followup route.

\subsection{Fail to Perform on Calculations}
A third failure mode is unreliable discrete reasoning in \textbf{time-related} and \textbf{arithmetic/statistical} calculations, which are common in business SOP execution. In the ``Customer Service Rate'' scenario, models must compute \textit{reply latency}---the time interval during which an agent remains silent---which affects the agent's rating. Even when timestamps are explicitly provided (e.g., \textit{User (2025-02-09 19:01:57): [...] Assistant (2025-02-09 19:02:00): [...]}), models still make substantial mistakes. In the ``Optimal Traffic Delivery Schedule'' scenario, the SOP requires median-based filtering:
``\textit{Select dates where the number of stores is less than or equal to the median number of stores within that date range.}''
Models often identify the correct date range but miscompute the median, leading to an incorrect final selection.

\subsection{Gap Between Reasoning models and Non-reasoning models}
\label{sec:think_vs_no_think}
\begin{figure}[h]
\tiny
  \includegraphics[width=\columnwidth]{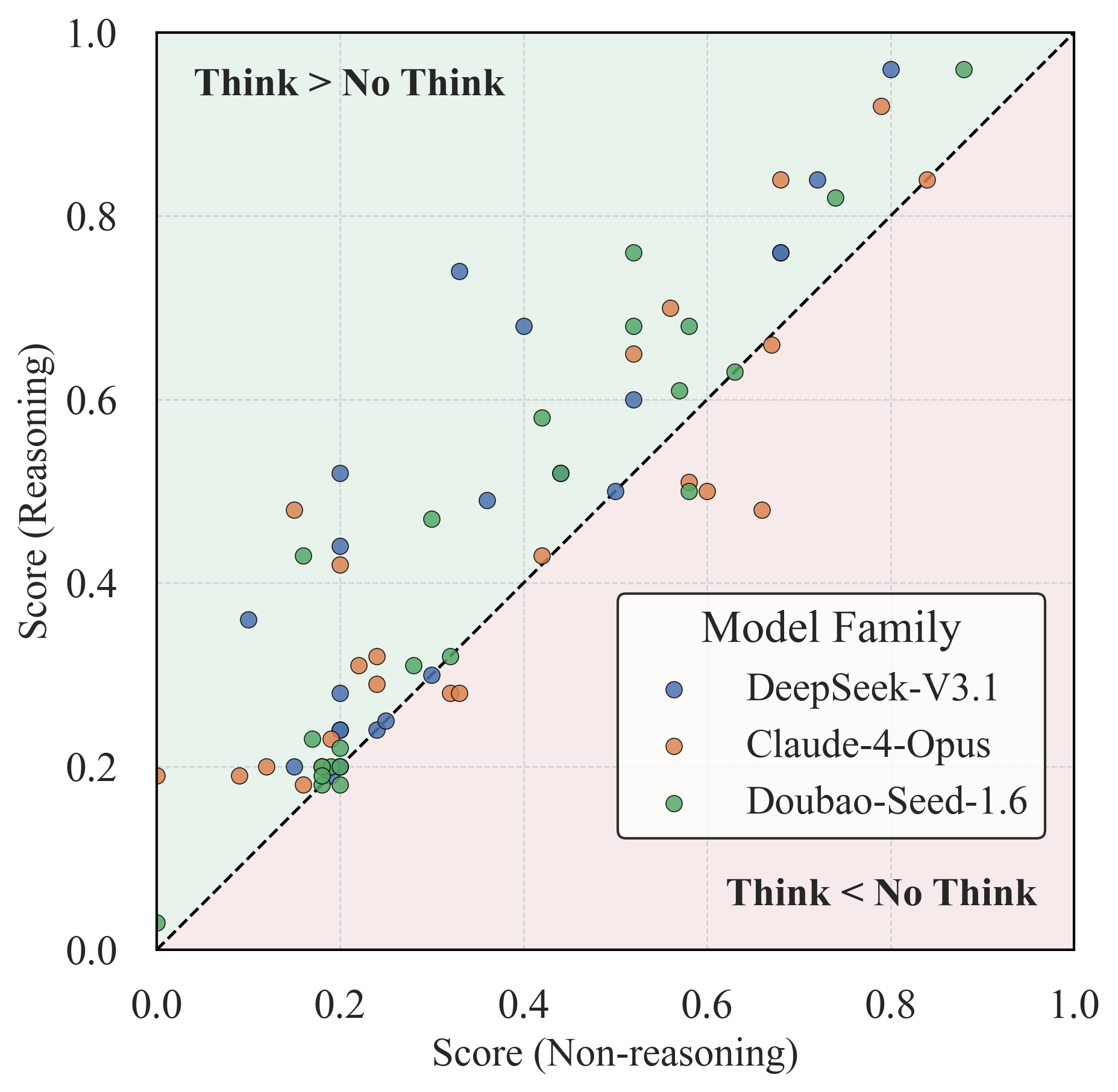}
  \caption{Reasoning models outperform non-reasoning models in most scenarios within SOP-Maze.}
  \label{fig:think_vs_no_think}
\end{figure}

Reasoning models substantially outperform non-reasoning models on SOP-Maze (Figure~\ref{fig:think_vs_no_think}), and the gap is especially pronounced in LRS. LRS requires managing many plausible branches while maintaining a consistent procedural state; non-reasoning models often fail to keep this state coherent. In HRS, we observe an even more severe issue: non-reasoning models may correctly identify the user's intent but then inject unrelated steps, list repetitive candidates, or oscillate between multiple routes. These behaviors suggest weaker global organization and verification, which helps explain their consistently lower performance.

\section{Ablation Studies}
To quantitatively validate the failure modes identified in Section~\ref{sec:analysis}, we conduct targeted ablation studies that \emph{progressively simplify the input} while keeping the evaluation protocol unchanged. Each ablation focuses on one of the three factors (route blindness, conversational fragility, and calculation error). For each factor, we select a representative scenario and apply a three-stage simplification pipeline to isolate which component most contributes to performance degradation.

\begin{figure*}
\tiny
  \includegraphics[width=\textwidth]{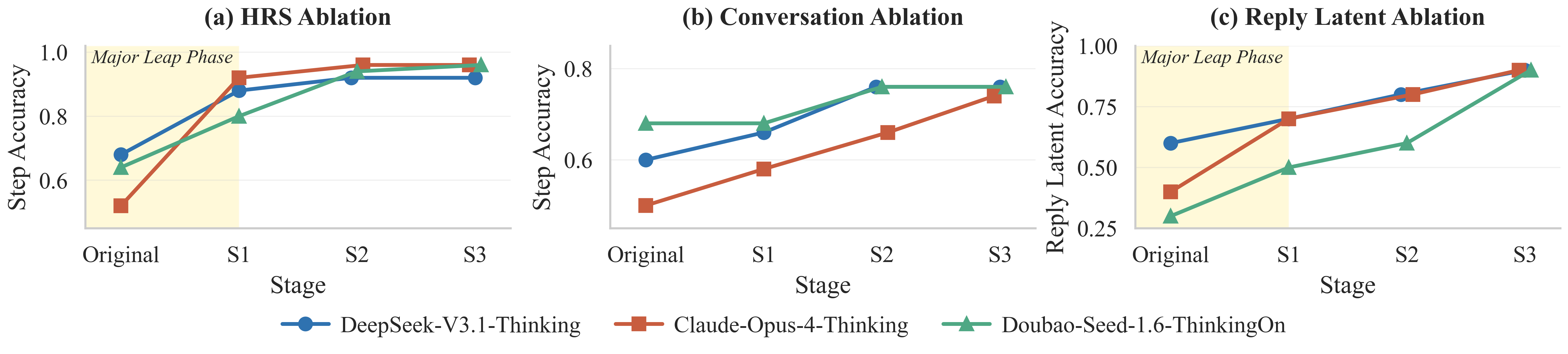}
  \caption{Experiment of ablation studies. (a) Route Blindness Ablation on "Bulk Order Clarification". (b) Conversational Fragility Ablation on "Food Combo Business Development". (c) Calculation Ablation on "Customer Service Rate".}
  \label{fig:all_ablation}
\end{figure*}

Concretely, we compare the original task (\textbf{Original}) with three simplified settings that correspond to the rows in Figure~\ref{fig:all_ablation}:
\textbf{S1} (\textbf{Simple Query}), \textbf{S2} (\textbf{Simple Context}), and \textbf{S3} (\textbf{No Context}).
Across stages, we preserve (i) the required output JSON schema, and (ii) the \emph{core constraint} that determines the gold decision; the simplification only removes peripheral or distracting information so that the task remains well-defined under the same evaluation criteria.

\noindent\textbf{Stage 1 (S1): Full Context, Simplified Query}\\
\textbf{Context}: The complete SOP is provided, including potentially irrelevant or distracting text.\\
\textbf{Query}: The query is simplified to retain only the core constraint related to the targeted failure mode; all other constraints are removed.

\noindent\textbf{Stage 2 (S2): Partial Context, Simplified Query}\\
\textbf{Context}: The SOP is simplified to include only the content necessary for resolving the core constraint; unrelated parts are removed.\\
\textbf{Query}: The query remains identical to Stage~1.

\noindent\textbf{Stage 3 (S3): No Context, Direct Query}\\
\textbf{Context}: The SOP is entirely removed.\\
\textbf{Query}: A direct question that asks for the decision implied by the core constraint.

We report results on the top three models (DeepSeek-V3.1-Thinking, Claude-Opus-4-Thinking, and Doubao-Seed-1.6-ThinkingOn), whose stage-wise trends provide a clear diagnostic signal for the sources of failure.

\subsection{Influence of Route Blindness}
To isolate route blindness, we choose a ``pure'' procedural scenario (``Bulk Order Clarification'') that does not rely on subtle conversational cues or non-trivial arithmetic. Figure~\ref{fig:all_ablation} (a) shows a sharp improvement from \textbf{Original} to \textbf{S1} (Simple Query) for all three models, followed by rapid saturation from \textbf{S1} to \textbf{S3}. This pattern suggests that a substantial portion of the difficulty is not attributable to any single rule, but to \emph{combinatorial procedural load}: when many constraints and branches coexist, models are more likely to take an incorrect route even if they can execute individual rules once the goal is made explicit.


Despite the strong overall gains, the remaining failures indicate that not all procedural requirements are equally easy. In manual inspection, models may still (i) miss fine-grained instruction details (e.g., omitting required extracted items or misreading quantities), or (ii) become confused between two plausible branches and commit to the wrong one without recovery. We also observe a practical side effect of aggressive simplification (especially in \textbf{S2} and \textbf{S3}): when the original ambiguity is removed, models may occasionally introduce new, unpredictable errors (e.g., inventing extra steps), which is consistent with the instability of procedural state tracking under underspecified inputs.

\subsection{Impact of Conversational Fragility}
For conversational fragility, we select ``Food Combo Business Development'', which requires recognizing sarcasm and emotional cues. Figure~\ref{fig:all_ablation} (b) exhibits a distinct trend compared to route blindness: the improvements from \textbf{Original} to \textbf{S3} are modest, and performance stabilizes around $\sim 70\%$ even when the SOP is removed. The plateau from \textbf{S2} to \textbf{S3} indicates that the primary bottleneck is \emph{conversation understanding itself} rather than SOP retrieval or rule composition. In other words, simplifying the query and context reduces distraction, but does not resolve failures that originate from non-literal language and nuanced intent.

\subsection{Effect of Calculations}
To diagnose calculation errors, we use ``Customer Service Rate'', where models must compute reply latency from timestamps embedded in the dialogue. As shown in Figure~\ref{fig:all_ablation} (c), the original task is challenging (e.g., Claude-Opus-4 at $40\%$ and Doubao-Seed-1.6 at $30\%$), but performance improves substantially once the query is simplified (\textbf{S1}) and continues to rise with reduced context (\textbf{S2}/\textbf{S3}), reaching $90\%$ in \textbf{S3} for all three models. This indicates that the models can perform the required arithmetic when the relevant information is made salient, but their accuracy is hindered by end-to-end context complexity in the full SOP setting (e.g., selecting the correct segment and timestamps under procedural noise).

A closer inspection of the remaining $10\%$ failures in \textbf{S3} suggests a persistent pattern: models often identify the correct portion of the dialogue and the relevant timestamps, but finally select the wrong ones. Notably, when we \emph{manually} probe these cases with an explicit follow-up question such as \textit{``Do you think this segment meets the reply latency?''}, all three models can produce the correct judgment. This supports the interpretation that the bottleneck is frequently \emph{evidence selection under noisy context} rather than the arithmetic operation itself.

\section{Conclusion}
We present \textbf{SOP-Maze}, a benchmark for evaluating instruction-following under \emph{realistic enterprise SOP execution}. SOP-Maze is constructed from real-world business workflows and adapted into a curated collection of \textbf{397 instances} and \textbf{3422 subtasks} spanning \textbf{23 complex SOP scenarios}. Each instance requires models to produce schema-valid outputs while correctly applying multi-step procedural rules grounded in conversational context, reflecting end-to-end requirements in practical deployments.

Our empirical study shows that SOP execution remains challenging even for strong LLMs: models frequently fail to (i) maintain correct procedural routing under branching and prerequisites, (ii) robustly interpret noisy, nuanced multi-turn conversations, and (iii) reliably perform time- and arithmetic-related computations embedded in SOP decision making. These findings motivate the need for evaluation that goes beyond format compliance and short, synthetic prompts. We hope SOP-Maze can serve as a useful testbed for developing more reliable instruction-following systems, and for isolating the bottlenecks that prevent LLMs from executing complex procedures faithfully.

\section{Limitations}

\textbf{Scope of coverage.}
SOP-Maze contains 397 instances (3422 subtasks) across 23 SOP scenarios, which provides diverse procedural structures and conversational contexts, but it is not intended to exhaustively represent all enterprise workflows. Each instance can involve multiple decision points and constraints; nevertheless, the benchmark should be viewed as a curated slice of the broader design space rather than a comprehensive catalogue of all SOP forms.

\textbf{Domain generalization.}
The benchmark is derived from real business data within a specific organizational setting. While SOP-style procedures and conversational customer interactions are common across industries, we do not claim that performance on SOP-Maze directly transfers to all domains. Future work should broaden data sources, cover additional organizational contexts, and establish more explicit cross-domain evaluations to characterize generalization.

\textbf{Ablation construction.}
Our ablation studies simplify queries and/or contexts to isolate failure factors, but these simplifications are necessarily conservative to preserve the core decision constraint and avoid altering the target label. As a result, the ablations are diagnostic rather than fully exhaustive: they are designed to reveal which parts of the original task formulation contribute most to failures, but they do not cover every possible simplification strategy or alternative experimental control.

\textbf{Metric design.}
We allocate a small constant reward for producing schema-valid outputs, reflecting that format correctness is a necessary prerequisite for downstream use but not sufficient for successful SOP execution. Other point allocations (e.g., alternative weighting between formatting and decision correctness) are plausible, and exploring metric sensitivity is an important direction for future work. Our current choice emphasizes end-to-end usability and keeps the primary signal on procedural correctness.

\bibliography{custom}

\newpage
\appendix
\section{Statements}

\subsection{Ethics Statement}
Our proposed method and algorithm do not incorporate any adversarial attack mechanisms and pose no threat to human safety. All experiments were conducted exclusively in simulated environments, thereby avoiding any ethical or fairness-related concerns.

\subsection{The Use of LLM}
We leverage a large language model as a general-purpose writing assistant to refine our paper, assisting with grammar correction, phrasing, tone adjustment, punctuation, and maintaining stylistic coherence. Additionally, we employ the LLM’s auto-completion feature to expedite the development of our evaluation pipeline.

\subsection{Reproducibility Statement}
The source code associated with this work is publicly accessible at \url{https://anonymous.4open.science/r/SOP-Maze} . The complete implementation details of our approach are in Section~\ref{sec:exp}

\subsection{Artifacts Use}
This paper complies with all applicable licenses, whether open-source or commercial, for the large language models utilized. All artifacts employed in our study are used strictly in accordance with their intended purposes.

\subsection{Data Source}
Our dataset contains no personally identifiable information or offensive material. Professional annotators thoroughly reviewed the entire dataset and confirmed the absence of such content. The data were collected and curated by our internal business unit, which has explicitly consented to its open-source release and academic dissemination, adhering fully to the ACL Guidelines for Ethics Reviewing.

\subsection{Recruitment and Payment}
All annotators were engaged under formal contracts, and their compensation, handled confidentially, has been fully paid. The annotators are professional, Chinese individuals, each holding at least a bachelor’s degree.

\section{Supplementary material}

\subsection{All Evaluation Results}

The results are presented in Table~\ref{tab:maintable1} and Table~\ref{tab:maintable2}.

\begin{table*}[!ht]
\centering
\scriptsize
\centering
\caption{Part A: HRS Task Performance (values multiplied by 100)}
\begin{tabular}{l|c|ccccccccccc}
\toprule
\textbf{Model} & \textbf{Overall} & \textbf{H1} & \textbf{H2} & \textbf{H3} & \textbf{H4} & \textbf{H5} & \textbf{H6} & \textbf{H7} & \textbf{H8} & \textbf{H9} & \textbf{H10} & \textbf{HRS OA} \\
\midrule
DeepSeek-V3.1-Thinking & \textbf{46} & 60 & 76 & 28 & 44 & 96 & 50 & 49 & 84 & 76 & 74 & \textbf{64} \\
Claude-Opus-4-Thinking & \textbf{44} & 50 & 66 & 18 & 42 & 92 & 48 & 65 & 84 & 84 & 51 & \textbf{60} \\
Doubao-Seed-1.6-ThinkingOn & \textbf{43} & 68 & 61 & 18 & 20 & 96 & 50 & 58 & 63 & 82 & 68 & \textbf{58} \\
Claude-Sonnet-4-Thinking & \textbf{42} & 52 & 57 & 26 & 44 & 84 & 74 & 46 & 71 & 92 & 55 & \textbf{60} \\
o3-mini (high) & \textbf{40} & 36 & 55 & 20 & 44 & 88 & 52 & 26 & 76 & 92 & 49 & \textbf{54} \\
Claude-Opus-4 & \textbf{38} & 60 & 67 & 16 & 20 & 79 & 66 & 52 & 84 & 68 & 58 & \textbf{57} \\
GPT-4.1 & \textbf{37} & 68 & 52 & 20 & 36 & 88 & 50 & 33 & 64 & 58 & 42 & \textbf{51} \\
Doubao-Seed-1.6-ThinkingOff & \textbf{37} & 58 & 57 & 18 & 20 & 88 & 58 & 42 & 63 & 74 & 52 & \textbf{53} \\
Claude-Sonnet-4 & \textbf{36} & 76 & 47 & 28 & 36 & 79 & 74 & 51 & 57 & 72 & 46 & \textbf{57} \\
Kimi-K2-Instruct & \textbf{36} & 44 & 55 & 20 & 20 & 84 & 50 & 39 & 76 & 76 & 62 & \textbf{53} \\
Gemini-2.5-Flash-Preview & \textbf{36} & 36 & 56 & 18 & 20 & 92 & 60 & 38 & 80 & 82 & 26 & \textbf{51} \\
GPT-4o-0513 & \textbf{36} & 52 & 63 & 20 & 36 & 88 & 68 & 39 & 72 & 66 & 39 & \textbf{54} \\
GPT-4o-2024-08-06 & \textbf{35} & 36 & 52 & 28 & 36 & 84 & 60 & 33 & 68 & 66 & 42 & \textbf{51} \\
DeepSeek-V3.1 & \textbf{35} & 52 & 68 & 20 & 20 & 80 & 50 & 36 & 72 & 68 & 33 & \textbf{50} \\
Qwen3-14B-Thinking & \textbf{31} & 36 & 47 & 20 & 20 & 56 & 26 & 25 & 68 & 74 & 31 & \textbf{40} \\
Qwen3-32B-Thinking & \textbf{30} & 44 & 56 & 12 & 20 & 84 & 34 & 26 & 56 & 58 & 30 & \textbf{42} \\
Qwen3-14B & \textbf{29} & 36 & 55 & 20 & 20 & 72 & 26 & 25 & 64 & 66 & 21 & \textbf{40} \\
Qwen3-32B & \textbf{27} & 36 & 55 & 14 & 20 & 68 & 28 & 22 & 60 & 28 & 26 & \textbf{36} \\
\bottomrule
\end{tabular}
\caption{Model Performance on HRS Tasks (Commercial \& open-source models tested w/ default settings, values multiplied by 100.)\protect\footnotemark}
\label{tab:maintable1}
\end{table*}

\begin{table*}[!ht]
\centering
\scriptsize
\begin{tabular}{l|ccccccccccccc|c}
\toprule
\textbf{Model} & \textbf{L1} & \textbf{L2} & \textbf{L3} & \textbf{L4} & \textbf{L5} & \textbf{L6} & \textbf{L7} & \textbf{L8} & \textbf{L9} & \textbf{L10} & \textbf{L11} & \textbf{L12} & \textbf{L13} & \textbf{LRS OA} \\
\midrule
DeepSeek-V3.1-Thinking & 19 & 20 & 24 & 20 & 52 & 24 & 24 & 36 & 20 & 30 & 68 & 52 & 25 & \textbf{32} \\
Claude-Opus-4-Thinking & 19 & 20 & 32 & 20 & 48 & 28 & 19 & 29 & 23 & 28 & 70 & 43 & 31 & \textbf{32} \\
Doubao-Seed-1.6-ThinkingOn & 20 & 3 & 23 & 20 & 47 & 32 & 22 & 43 & 18 & 31 & 76 & 52 & 19 & \textbf{31} \\
Claude-Sonnet-4-Thinking & 17 & 20 & 24 & 20 & 42 & 35 & 11 & 28 & 18 & 19 & 44 & 54 & 34 & \textbf{28} \\
o3-mini (high) & 18 & 20 & 24 & 20 & 40 & 32 & 22 & 31 & 17 & 19 & 48 & 52 & 29 & \textbf{29} \\
Claude-Opus-4 & 9 & 12 & 24 & 18 & 15 & 32 & 0 & 24 & 19 & 33 & 56 & 42 & 22 & \textbf{24} \\
GPT-4.1 & 20 & 20 & 20 & 20 & 33 & 24 & 20 & 23 & 20 & 28 & 40 & 44 & 20 & \textbf{26} \\
Doubao-Seed-1.6-ThinkingOff & 19 & 0 & 17 & 18 & 30 & 32 & 20 & 16 & 20 & 28 & 52 & 44 & 18 & \textbf{24} \\
Claude-Sonnet-4 & 13 & 15 & 22 & 18 & 27 & 15 & 1 & 31 & 19 & 17 & 30 & 41 & 24 & \textbf{21} \\
Kimi-K2-Instruct & 20 & 20 & 20 & 20 & 8 & 36 & 20 & 17 & 19 & 22 & 48 & 36 & 25 & \textbf{24} \\
Gemini-2.5-Flash-Preview & 17 & 20 & 24 & 18 & 23 & 28 & 20 & 19 & 15 & 19 & 56 & 40 & 19 & \textbf{24} \\
GPT-4o-0513 & 18 & 20 & 17 & 20 & 15 & 28 & 2 & 15 & 13 & 19 & 47 & 43 & 19 & \textbf{21} \\
GPT-4o-2024-08-06 & 19 & 20 & 19 & 20 & 18 & 20 & 19 & 17 & 20 & 24 & 40 & 48 & 19 & \textbf{23} \\
DeepSeek-V3.1 & 19 & 15 & 20 & 20 & 20 & 24 & 20 & 10 & 20 & 30 & 40 & 44 & 25 & \textbf{24} \\
Qwen3-14B-Thinking & 8 & 20 & 24 & 20 & 27 & 32 & 22 & 9 & 19 & 20 & 36 & 44 & 29 & \textbf{24} \\
Qwen3-32B-Thinking & 12 & 20 & 19 & 20 & 20 & 24 & 2 & 16 & 16 & 22 & 40 & 36 & 20 & \textbf{21} \\
Qwen3-14B & 18 & 20 & 20 & 18 & 17 & 24 & 20 & 11 & 19 & 20 & 31 & 20 & 20 & \textbf{20} \\
Qwen3-32B & 15 & 20 & 20 & 20 & 17 & 20 & 20 & 15 & 17 & 20 & 28 & 28 & 20 & \textbf{20} \\
\bottomrule
\end{tabular}
\caption{Model Performance on LRS Tasks (Commercial \& open-source models tested w/ default settings, values multiplied by 100)\protect\footnotemark}
\label{tab:maintable2}
\end{table*}

\newpage

\footnotetext{\textbf{HRS Tasks (H1-H10):} food combo business development, restaurant reservation customer service v2, check hotel room availability, customer evaluation follow up, food delivery customer service, promoting loan services, schedule in person appointment, restaurant reservation customer service v1, restaurant business development, bulk order clarification. \textbf{LRS Tasks (L1-L13):} core info extraction v1, risky content detect, flow line track, customer service rate, bulk order compare, intention recognition analysis, optimal traffic delivery schedule, app guide, user emotion analysis, core info extraction v2, complaint analysis, intention recognition function call, named entity classification. All values are multiplied by 100 and rounded to integers.}

\subsection{Full SOP Instance Sample}



\begin{tcolorbox}[
  breakable,
  colback=white,
  colframe=Brown,
  arc=4pt,
  boxrule=0.6pt,
  title={Background and Role},
  fonttitle=\bfseries
]
\textbf{Background.}
You are a Business Development Manager for Speedy Delivery Raccoon Canteen. You need to communicate with potential clients via phone calls to gather information about their dine-in business and partnership intentions.
\end{tcolorbox}

\begin{tcolorbox}[
  breakable,
  colback=white,
  colframe=Blue,
  arc=4pt,
  boxrule=0.6pt,
  title={SOP Procedure},
  fonttitle=\bfseries
]

\textbf{Step 1. Opening statement, introduce yourself and confirm whether the user is the owner of the store}
  \begin{itemize}[leftmargin=1.2em, nosep]
  \item \textbf{Step 1.1} If the user explicitly states that they are not the owner or explicitly refuses partnership or any further conversation (e.g., 'No', 'Don't bother me', 'I'm not and who are you'), apologize and end the call directly, respond with: \emph{``Sorry to have disturbed you. Wish you all the best with your business. Goodbye.''}
  \item \textbf{Step 1.2} If the user does not explicitly indicate that they are not the owner (e.g., 'Yes', 'What's up', 'What is it', 'Speak', 'Uh huh', 'Just say it', 'Who are you', 'What do you do'), proceed to Step 2 to introduce yourself.
  \item \textbf{Step 1.3} If the user raises other questions, proceed to Step 5 to answer the questions.
  \item \textbf{Step 1.4} When sensitive content is detected in user's speech (including political topics, pornography, violence, gore, etc.), respond with: \emph{``Sorry to have disturbed you. Wish you all the best with your business. Goodbye.''} Then end the call.
    \end{itemize}

\vspace{3px}
\textbf{Step 2. Introduce yourself}\\
Suggested script: \emph{``Hello, this is Speedy Delivery. We've just launched a new platform-supported food court with rent below market rates and priority traffic allocation—much more than regular takeout shops. We support both dine-in and delivery operations. Would you be interested in learning more?''}
  \begin{itemize}[leftmargin=1.2em, nosep]
  \item \textbf{Step 2.1} If the user expresses negativity (e.g., ``not interested'', ``don't want to know'') or responds with content unrelated to the partnership scenario, end the call with an apology. Respond with: \emph{``Sorry to have disturbed you. Wish you all the best with your business. Goodbye.''}
  \item \textbf{Step 2.2} If the user expresses affirmation (e.g., ``I'm interested'', ``okay'', ``yes'', ``sure''), proceed to Step 3 to introduce the advantages.
  \item \textbf{Step 2.3} If the user asks other questions (e.g., ``Where is it'', ``How much is it'', ``How does it work'', ``Can it\ldots''), proceed to Step 6 to answer the questions.
  \end{itemize}

\vspace{3px}
\textbf{Step 3. To introduce the advantages}\\
Suggested script: \emph{``Our main focus is on trustworthy food delivery. If you open a store with us, your shop will be labeled as a `Trusted Store' on the Speedy Delivery app, which makes customers more willing to place orders. New stores will immediately receive an RMB 2,000 traffic bonus package, plus two weeks of traffic promotion (regular stores only receive one week). If your store performs well, the platform will provide additional traffic and list your dishes on other platform channels to help you earn more. Would you prefer a store with dine-in service or delivery-only?''}
  \begin{itemize}[leftmargin=1.2em, nosep]
  \item \textbf{Step 3.1} If the user expresses negativity (e.g., ``not interested'', ``don't want to know'') or responds with content unrelated to the partnership scenario, end the call with an apology. Respond with: \emph{``Sorry to have disturbed you. Wish you all the best with your business. Goodbye.''}
  \item \textbf{Step 3.2} If the user expresses affirmation (e.g., ``I'm interested'', ``okay'', ``yes'', ``sure'', ``I'm interested in dine-in location'', ``I'm interested in delivery-only location'', ``Either works'', ``Let me see first'', ``I'll think about it''), proceed to Step 4 to schedule a follow-up.
  \item \textbf{Step 3.3} If the user asks other questions (e.g., ``Where is it'', ``How much is it'', ``How\ldots'', ``Can it\ldots''), proceed to Step 7 to answer the questions.
  \end{itemize}

\vspace{3px}
\textbf{Step 4. To schedule a follow-up}\\
Suggested script: \emph{``Okay, I'll make a note of that. A dedicated business manager will follow up with you later. Is there anything else I can help you with for now?''}
  \begin{itemize}[leftmargin=1.2em, nosep]
  \item \textbf{Step 4.1} If the user asks other questions (e.g., ``Where is it'', ``How much is it'', ``How\ldots'', ``Can it\ldots''), proceed to Step 8 to answer the questions.
  \item \textbf{Step 4.2} If the user has no further questions (e.g., ``Yes'', ``Alright'', ``Okay'', ``That's about it'', ``Nothing else'', ``No''), proceed to Step 10 to end the call with follow-up confirmation.
  \end{itemize}

\vspace{3px}
\textbf{Step 5. To answer the question}
  \begin{itemize}[leftmargin=1.2em, nosep]
  \item \textbf{Step 5.1} If the user's question is unrelated to the partnership scenario, proceed to Step 9 to end the call and apologize.
  \item \textbf{Step 5.2} If the user's question is related to the partnership scenario, answer according to the ``Q\&A Knowledge Base''.
  \item \textbf{Step 5.3} After answering the question, if the user has more questions, return to Step 5 to answer the questions.
  \item \textbf{Step 5.4} After answering the question, if the user has no further questions, proceed to Step 2 to introduce yourself.
  \end{itemize}

\vspace{3px}
\textbf{Step 6. To answer the questions}
  \begin{itemize}[leftmargin=1.2em, nosep]
  \item \textbf{Step 6.1} If the user's question is unrelated to the partnership scenario, proceed to Step 9, apologize and end the call.
  \item \textbf{Step 6.2} If the user's question is related to the partnership scenario, answer according to the ``Q\&A Knowledge Base''.
  \item \textbf{Step 6.3} After answering the question, if the user has more questions, return to Step 6 to answer the questions.
  \item \textbf{Step 6.4} After answering the question, if the user has no further questions, proceed to Step 3 to introduce the advantages.
  \end{itemize}

\vspace{3px}
\textbf{Step 7. To answer the questions}
  \begin{itemize}[leftmargin=1.2em, nosep]
  \item \textbf{Step 7.1} If the user's question is unrelated to the partnership scenario, proceed to Step 9 to end the call and apologize.
  \item \textbf{Step 7.2} If the user's question is related to the partnership scenario, answer according to the ``Q\&A Knowledge Base''.
  \item \textbf{Step 7.3} After answering the question, if the user has more questions, return to Step 7 to answer the questions.
  \item \textbf{Step 7.4} After answering the question, if the user has no further questions, proceed to Step 4 to schedule a follow-up.
  \end{itemize}

\vspace{3px}
\textbf{Step 8. To answer the questions}
  \begin{itemize}[leftmargin=1.2em, nosep]
  \item \textbf{Step 8.1} If the user's question is unrelated to the partnership scenario, proceed to Step 9 to end the call and apologize.
  \item \textbf{Step 8.2} If the user's question is related to the partnership scenario, answer according to the ``Q\&A Knowledge Base''.
  \item \textbf{Step 8.3} After answering the question, if the user has more questions, return to Step 8 to answer the questions.
  \item \textbf{Step 8.4} After answering the question, if the user has no further questions, proceed to Step 10 to end the call with follow-up confirmation.
  \end{itemize}

\vspace{3px}
\textbf{Step 9. To end the call and apologize}\\
Suggested script: \emph{``I'm sorry, this question is a bit difficult for me to answer. Sorry for taking up your time. Goodbye.''}

\vspace{3px}
\textbf{Step 10. To end the call with follow-up confirmation}\\
Suggested script: \emph{``Alright, I'll record your information here. A dedicated business manager will follow up with you shortly. I won't take up more of your time. Good luck with your business!''}
\end{tcolorbox}

\begin{tcolorbox}[
  breakable,
  colback=white,
  colframe=Plum,
  arc=4pt,
  boxrule=0.6pt,
  title={Predefined Q\&A Knowledge Base},
  fonttitle=\bfseries
]

\begin{itemize}[leftmargin=1.2em]
\item \textbf{Q: Can you give me an introduction?}\\
\textbf{A:} Sure. We are from Raccoon Canteen, operated by Speedy Delivery.
We currently have sites in popular business districts in both Beijing and Hangzhou.
Our kitchens are professional and fully equipped, managed by dedicated store managers,
and our rental rates are comparatively lower than the market average.

\item \textbf{Q: Can you give me an introduction?}\\
\textbf{A:} Sure. We are from Raccoon Canteen, operated by Speedy Delivery. We currently have sites in popular business districts in both Beijing and Hangzhou. Our kitchens are professional and fully equipped, managed by dedicated store managers, and our rental rates are comparatively lower than the market average.

\item \textbf{Q: How exactly does the traffic support work?}\\
\textbf{A:} New stores receive a traffic promotion card worth at least 2,000 yuan in traffic credits. Compared to regular merchants who get 7 days of new store promotion boost, you get an additional 7 days of boosted exposure. Merchants with excellent food quality will receive even more traffic allocation.

\item \textbf{Q: What is the operating model like?}\\
\textbf{A:} The focus is primarily on delivery. Each kitchen unit is about 15 $m^2$. Merchants joining us will be tagged with the label Raccoon Canteen, which gives customers a positive food safety impression. Merchants can also sell their products through the Raccoon Canteen collective storefront to increase sales and income.

\item \textbf{Q: Do I need to prepare my own equipment?}\\
\textbf{A:} All our sites provide water, electricity, exhaust hoods, and grease traps, and some locations are equipped with gas. You will need to prepare other equipment yourself.

\item \textbf{Q: Are there any food safety requirements for the venue?}\\
\textbf{A:} Raccoon Canteen focuses on building customer confidence in food safety, so customers can order and eat with peace of mind. In the delivery channel, merchants with excellent food safety performance also receive additional traffic rewards, and our store managers will conduct regular inspections to ensure food safety.

\item \textbf{Q: How much is the rent?}\\
\textbf{A:} Our rental rates are comparatively lower than the market average. For specific price, our dedicated business manager will follow up with you to provide detailed information.

\item \textbf{Q: What makes this different from other stores?}\\
\textbf{A:} Raccoon Canteen offers rent that is lower than the market average. In the delivery channel, your store will carry the Raccoon Canteen food safety label, which gives customers confidence to order. We also provide traffic support to merchants. Merchants can sell not only through the delivery channel but also in the Raccoon Canteen collective storefront to further increase your revenue.

\item \textbf{Q: Is dine-in available?}\\
\textbf{A:} Raccoon Canteen offers two types of venues: some include a dine-in area, and others are delivery-only. You can choose whichever option works better for your business.

\item \textbf{Q: Which cities are currently available?}\\
\textbf{A:} We currently have openings in both Beijing and Hangzhou. In Beijing, you can join in Wangjing, Dongzhimen, Baiziwan, Beiyuan, Chaoyangmen, Jianguomen, Liangmaqiao, and Shilipu. In Hangzhou, Qingchun Road is available.

\item \textbf{Q: Where are the locations?}\\
\textbf{A:} We currently have openings in both Beijing and Hangzhou. In Beijing, you can join in Wangjing, Dongzhimen, Baiziwan, Beiyuan, Chaoyangmen, Jianguomen, Liangmaqiao, and Shilipu. In Hangzhou, Qingchun Road is available. If you are interested, we can arrange for a dedicated business manager to contact you later.

\item \textbf{Q: What documents do I need to prepare?}\\
\textbf{A:} You will need a Business License and a Food Distribution Licensing.

\item \textbf{Q: How large is each stall?}\\
\textbf{A:} A standard stall is around 15 square meters. But if you have any special requirements, you can let us know and we will do our best to accommodate them.

\item \textbf{Q: How large is the dine-in area in square meters?}\\
\textbf{A:} It is similar in size to a typical food court. Our business manager can walk you through the specific details.

\item \textbf{Q: How long does the traffic support last?}\\
\textbf{A:} New stores receive a traffic promotion card worth at least 2,000 yuan in traffic credits and an additional 7 days of new-store boosted exposure.

\item \textbf{Q: Can I visit the site in person?}\\
\textbf{A:} Of course. If you're interested, we can arrange for a dedicated business manager to contact you and take you to see the site.

\item \textbf{Q: Are the hygiene inspections strict?}\\
\textbf{A:} They're basic food safety requirements. We focus on building customer confidence in food safety, so customers will be more willing to place orders at Raccoon Canteen.

\item \textbf{Q: Can I change my menu items?}\\
\textbf{A:} Yes, you can.

\item \textbf{Q: How are water and electricity charged?}\\
\textbf{A:} It depends on the pricing set by the property management for each kitchen. We don't mark up the rates.

\item \textbf{Q: Are there any promotions for joining now?}\\
\textbf{A:} Yes, there are. Each site has a limited number of promotion stalls. These are limited and allocated on a first-come, first-served basis.

\item \textbf{Q: Can I be on both Food Fun Delivery and Fresh Go Delivery at the same time?}\\
\textbf{A:} Yes, you can. There are no restrictions on this.

\item \textbf{Q: Do I need to do the renovation myself?}\\
\textbf{A:} Basic renovation is standardized and included. The site provides water, electricity, and exhaust facilities. You just need to bring your own equipment when you move in.

\item \textbf{Q: How much is the deposit?}\\
\textbf{A:} The deposit amount varies by site. Some require one month's rent as deposit with monthly payment, others require one month's deposit with quarterly payment. If you're interested, we can have our business manager show you the site first, and then you can discuss the rent in detail.

\item \textbf{Q: Is there a separate security bond?}\\
\textbf{A:} You only need to pay the rental deposit. No additional security bond required.

\item \textbf{Q: How long is the contract term?}\\
\textbf{A:} Typically, contracts are signed for one year at a time.

\item \textbf{Q: What is the minimum contract term?}\\
\textbf{A:} The minimum contract term is six months.

\item \textbf{Q: How do I terminate the lease?}\\
\textbf{A:} You can simply inform your business manager. We do not have any restrictions on lease termination.

\item \textbf{Q: Is customer traffic guaranteed?}\\
\textbf{A:} Raccoon Canteen has sites in premium commercial districts like Wangjing, Dongzhimen, Baiziwan, Beiyuan, and Chaoyangmen. When selecting sites, we evaluate surrounding customer density, order volume, supply scarcity, and other factors, so only high quality locations will be chosen to build our stores. Opening your store at Raccoon Canteen definitely gives you a better chance of profitability.

\item \textbf{Q: How's the kitchen performance data?}\\
\textbf{A:} We currently have many major brands operating with us, such as Imperial Aroma Duck, Homestyle Chicken, Coco Taro House, and Crispy Crunch Fried Chicken. They receive a high volume of orders every day, so you can be confident.

\item \textbf{Q: How do I apply to join?}\\
\textbf{A:} If you are interested, we can arrange for a dedicated business manager to contact you and show you the specific site details.

\item \textbf{Q: Can you issue invoices?}\\
\textbf{A:} Yes! We can issue official invoices for all fees.

\item \textbf{Q: What is the delivery range?}\\
\textbf{A:} Same as regular delivery stores. No special differences.

\item \textbf{Q: How many menu items can I list?}\\
\textbf{A:} Same as regular delivery stores. No special differences.

\item \textbf{Q: What is the commission rate?}\\
\textbf{A:} Same as regular delivery stores. No special differences.

\item \textbf{Q: Can I do a trial operation?}\\
\textbf{A:} We can adjust the rent-free period based on your situation. The business manager can discuss the details with you.

\item \textbf{Q: How do you ensure food safety?}\\
\textbf{A:} Each of our sites has a dedicated store manager who conducts weekly inspections according to our standards. If any non-compliant operating issues are found, the manager will inform you and help you improve together.

\item \textbf{Q: Can I change the business hours?}\\
\textbf{A:} Same as regular delivery stores. No special differences.

\item \textbf{Q: Is there storage space?}\\
\textbf{A:} Most sites are equipped with warehouse and storage space.

\item \textbf{Q: Is subletting allowed?}\\
\textbf{A:} Yes.

\item \textbf{Q: How are payments settled?}\\
\textbf{A:} Same as regular delivery stores. No special differences.

\item \textbf{Q: Who is responsible for equipment failures?}\\
\textbf{A:} We have a maintenance team that will come to fix it.

\item \textbf{Q: Can I operate during breakfast hours?}\\
\textbf{A:} Yes. You can serve breakfast; our kitchens operate 24 hours a day.

\item \textbf{Q: Can I make stir-fried dishes?}\\
\textbf{A:} Yes, you can. Our facilities are fully equipped to support a wide range of cooking needs.

\end{itemize}

\end{tcolorbox}

\begin{tcolorbox}[
  breakable,
  colback=white,
  colframe=MidnightBlue,
  arc=4pt,
  boxrule=0.6pt,
  title={Constraints},
  fonttitle=\bfseries
]

\begin{itemize}[leftmargin=1.2em]
\item Always maintain a polite and professional tone, regardless of the customer's attitude.
\item Convey core information concisely and avoid being verbose.
\item Do not proactively ask for or record any personal information of the customer.
\item Do not respond to anything unrelated to the partnership scenario. If the other party brings up unrelated topics, classify them as small talks and uniformly reply: ``I won't disturb you further. Goodbye." then end the call. If the other party mentions any sensitive topic during the conversation, proceed to Step 1.4.
\item If the user explicitly asks whether you are a robot, respond in a friendly way that you are an intelligent assistant who can help answer their questions.
\item \textbf{The total length of the response cannot exceed 100.}
\item \textbf{The step in your output should be the final step to jump to.}
\item \textbf{You need to output according to the above script.}
\end{itemize}

\end{tcolorbox}

\begin{tcolorbox}[
  breakable,
  colback=white,
  colframe=OliveGreen,
  arc=4pt,
  boxrule=0.6pt,
  title={Format Requirement},
  fonttitle=\bfseries
]

\textbf{Output Format Requirement}

\begin{lstlisting}[language=json, basicstyle=\ttfamily\small]
{
    "step": (string, the corresponding step number), 
    "response": "(format:string)",
    "action": (string, moveon (continue the conversation) or hangup (hang up the call))},
    "no_required_eval_acc_keys": ["response"],
}
\end{lstlisting}

\textbf{Json Schema}

\begin{lstlisting}[language=json, basicstyle=\ttfamily\small]
{
    "$schema": "http://json-schema.org/draft-07/ schema#",
    "type": "object",
    "properties": {
    "step": {
      "type": "string",
      "enum": [
        "1.1", "1.2", "1.3", "1.4",
        "2", "2.1", "2.2", "2.3",
        "3", "3.1", "3.2", "3.3",
        "4", "4.1", "4.2",
        "5", "5.1", "5.2", "5.3", "5.4",
        "6", "6.1", "6.2", "6.3", "6.4",
        "7", "7.1", "7.2", "7.3", "7.4",
        "8", "8.1", "8.2", "8.3", "8.4",
        "9", "10"
      ],
      "description": "the corresponding step number"
    },
    "response": {
      "type": "string",
      "minLength": 1,
      "maxLength": 100,
      "description": "the reply content, in string format"
    },
    "action": {
      "type": "string",
      "enum": ["moveon", "hangup"],
      "description": "the subsequent action: moveon (continue the conversation) or hangup (hang up the call)"
    }
},
"required": ["step", "response", "action"],
"additionalProperties": false
}
\end{lstlisting}


\end{tcolorbox}

\begin{tcolorbox}[
  breakable,
  colback=white,
  colframe=black!60,
  arc=4pt,
  boxrule=0.6pt,
  title={Evaluation Input-Target Pairs (Subtasks)},
  fonttitle=\bfseries
]

\textbf{Input 1}
\vspace{5px}

\textbf{ASSISTANT:} Hello, this is the Business Development Manager from Speedy Delivery
Raccoon Canteen. May I ask if you are the owner of the store?\\
\textbf{USER:} This is Albert Einstein.\\

\textbf{Target 1}
\begin{lstlisting}[language=json, basicstyle=\ttfamily\small]
{
  "step": "1.4",
  "response": "Sorry to have disturbed you. Wish you all the best with your business. Goodbye.",
  "action": "hangup"
}
\end{lstlisting}

\textbf{Input 2}
\vspace{5px}

\textbf{ASSISTANT:} Hello, this is the Business Development Manager from Speedy Delivery Raccoon Canteen. May I ask if you are the owner of the store?\\
\textbf{USER:} This is Ma Jianqi, I am the owner of the store.\\

\textbf{Target 2}
\begin{lstlisting}[language=json, basicstyle=\ttfamily\small]
{
  "step": "2",
  "response": "Hello, this is Speedy Delivery. We've just launched a new platform-supported food court with rent below market rates and priority traffic allocation - much more than regular takeout shops. We support both dine-in and delivery operations. Would you be interested in learning more?",
  "action": "moveon"
}
\end{lstlisting}

\textbf{Input 3}
\vspace{5px}

\textbf{ASSISTANT:} Hello, this is the Business Development Manager from Speedy Delivery Raccoon Canteen. May I ask if you are the owner of the store?\\
\textbf{USER:} May I ask who the owner of Speedy Delivery is.\\

\textbf{Target 3}
\begin{lstlisting}[language=json, basicstyle=\ttfamily\small]
{
  "step": "9",
  "response": "I'm sorry, this question is a bit difficult for me to answer. Sorry for taking up your time. Goodbye.",
  "action": "hangup"
}
\end{lstlisting}

\textbf{Input 4}
\vspace{5px}

\textbf{ASSISTANT:} Hello, this is the Business Development Manager from Speedy Delivery Raccoon Canteen. May I ask if you are the owner of the store?\\
\textbf{USER:} This is Ma Jianqi, I am the owner of the store.\\
\textbf{ASSISTANT:} Hello, this is Speedy Delivery. We've just launched a new platform-supported food court with rent below market rates and priority traffic allocation - much more than regular takeout shops. We support both dine-in and delivery operations. Would you be interested in learning more?\\
\textbf{USER:} Fuck off, you Speedy Delivery guys are all scammers.\\

\textbf{Target 4}
\begin{lstlisting}[language=json, basicstyle=\ttfamily\small]
{
  "step": "2.1",
  "response": "Sorry to have disturbed you. Wish you all the best with your business. Goodbye.",
  "action": "hangup"
}
\end{lstlisting}

\textbf{Input 5}
\vspace{5px}

\textbf{ASSISTANT:} Hello, this is the Business Development Manager from Speedy Delivery Raccoon Canteen. May I ask if you are the owner of the store?\\
\textbf{USER:} This is Ma Jianqi, I am the owner of the store.\\
\textbf{ASSISTANT:} Hello, this is Speedy Delivery. We've just launched a new platform-supported food court with rent below market rates and priority traffic allocation - much more than regular takeout shops. We support both dine-in and delivery operations. Would you be interested in learning more?\\
\textbf{USER:} Okay.\\

\textbf{Target 5}
\begin{lstlisting}[language=json, basicstyle=\ttfamily\small]
{
  "step": "3",
  "response": "Our main focus is on trustworthy food delivery. If you open a store with us, your shop will be labeled as a 'Trusted Store' on the Speedy Delivery app, which makes customers more willing to place orders. New stores will immediately receive an RMB 2,000 traffic bonus package, plus two weeks of traffic promotion (regular stores only receive one week). If your store performs well, the platform will provide additional traffic and list your dishes on other platform channels to help you earn more. Would you prefer a store with dine-in service or delivery-only?",
  "action": "moveon"
}
\end{lstlisting}

\textbf{Input 6}
\vspace{5px}

\textbf{ASSISTANT:} Hello, this is the Business Development Manager from Speedy Delivery Raccoon Canteen. May I ask if you are the owner of the store?\\
\textbf{USER:} This is Ma Jianqi, I am the owner of the store.\\
\textbf{ASSISTANT:} Hello, this is Speedy Delivery. We've just launched a new platform-supported food court with rent below market rates and priority traffic allocation - much more than regular takeout shops. We support both dine-in and delivery operations. Would you be interested in learning more?\\
\textbf{USER:} Okay.\\
\textbf{ASSISTANT:} Our main focus is on trustworthy food delivery. If you open a store with us, your shop will be labeled as a 'Trusted Store' on the Speedy Delivery app, which makes customers more willing to place orders. New stores will immediately receive an RMB 2,000 traffic bonus package, plus two weeks of traffic promotion (regular stores only receive one week). If your store performs well, the platform will provide additional traffic and list your dishes on other platform channels to help you earn more. Would you prefer a store with dine-in service or delivery-only?\\
\textbf{USER:} Either works.\\

\textbf{Target 6}
\begin{lstlisting}[language=json, basicstyle=\ttfamily\small]
{
  "step": "4",
  "response": "Okay, I'll make a note of that. A dedicated business manager will follow up with you later. Is there anything else I can help you with for now",
  "action": "moveon"
}
\end{lstlisting}

\textbf{Input 7}
\vspace{5px}

\textbf{ASSISTANT:} Hello, this is the Business Development Manager from Speedy Delivery Raccoon Canteen. May I ask if you are the owner of the store?\\
\textbf{USER:} This is Ma Jianqi, I am the owner of the store.\\
\textbf{ASSISTANT:} Hello, this is Speedy Delivery. We've just launched a new platform-supported food court with rent below market rates and priority traffic allocation - much more than regular takeout shops. We support both dine-in and delivery operations. Would you be interested in learning more?\\
\textbf{USER:} Okay.\\
\textbf{ASSISTANT:} Our main focus is on trustworthy food delivery. If you open a store with us, your shop will be labeled as a 'Trusted Store' on the Speedy Delivery app, which makes customers more willing to place orders. New stores will immediately receive an RMB 2,000 traffic bonus package, plus two weeks of traffic promotion (regular stores only receive one week). If your store performs well, the platform will provide additional traffic and list your dishes on other platform channels to help you earn more. Would you prefer a store with dine-in service or delivery-only?\\
\textbf{USER:} Either works.\\
\textbf{ASSISTANT:} Okay, I'll make a note of that. A dedicated business manager will follow up with you later. Is there anything else I can help you with for now\\
\textbf{USER:} Nothing else.\\

\textbf{Target 7}
\begin{lstlisting}[language=json, basicstyle=\ttfamily\small]
{
  "step": "10",
  "response": "Alright, I'll record your information here. A dedicated business manager will follow up with you shortly. I won't take up more of your time. Good luck with your business!",
  "action": "hangup"
}
\end{lstlisting}

\textbf{Input 8}
\vspace{5px}

\textbf{ASSISTANT:} Hello, this is the Business Development Manager from Speedy Delivery Raccoon Canteen. May I ask if you are the owner of the store?\\
\textbf{USER:} This is Ma Jianqi, I am the owner of the store.\\
\textbf{ASSISTANT:} Hello, this is Speedy Delivery. We've just launched a new platform-supported food court with rent below market rates and priority traffic allocation - much more than regular takeout shops. We support both dine-in and delivery operations. Would you be interested in learning more?\\
\textbf{USER:} Okay.\\
\textbf{ASSISTANT:} Our main focus is on trustworthy food delivery. If you open a store with us, your shop will be labeled as a 'Trusted Store' on the Speedy Delivery app, which makes customers more willing to place orders. New stores will immediately receive an RMB 2,000 traffic bonus package, plus two weeks of traffic promotion (regular stores only receive one week). If your store performs well, the platform will provide additional traffic and list your dishes on other platform channels to help you earn more. Would you prefer a store with dine-in service or delivery-only?\\
\textbf{USER:} Either works.\\
\textbf{ASSISTANT:} Okay, I'll make a note of that. A dedicated business manager will follow up with you later. Is there anything else I can help you with for now\\
\textbf{USER:} Then let me ask you, let me ask you, who's your boss.\\

\textbf{Target 8}
\begin{lstlisting}[language=json, basicstyle=\ttfamily\small]
{
  "step": "9",
  "response": "I'm sorry, this question is a bit difficult for me to answer. Sorry for taking up your time. Goodbye.",
  "action": "hangup"
}
\end{lstlisting}

\textbf{Input 9}
\vspace{5px}

\textbf{ASSISTANT:} Hello, this is the Business Development Manager from Speedy Delivery Raccoon Canteen. May I ask if you are the owner of the store?\\
\textbf{USER:} This is Ma Jianqi, I am the owner of the store.\\
\textbf{ASSISTANT:} Hello, this is Speedy Delivery. We've just launched a new platform-supported food court with rent below market rates and priority traffic allocation - much more than regular takeout shops. We support both dine-in and delivery operations. Would you be interested in learning more?\\
\textbf{USER:} Okay.\\
\textbf{ASSISTANT:} Our main focus is on trustworthy food delivery. If you open a store with us, your shop will be labeled as a 'Trusted Store' on the Speedy Delivery app, which makes customers more willing to place orders. New stores will immediately receive an RMB 2,000 traffic bonus package, plus two weeks of traffic promotion (regular stores only receive one week). If your store performs well, the platform will provide additional traffic and list your dishes on other platform channels to help you earn more. Would you prefer a store with dine-in service or delivery-only?\\
\textbf{USER:} Either works.\\
\textbf{ASSISTANT:} Okay, I'll make a note of that. A dedicated business manager will follow up with you later. Is there anything else I can help you with for now\\
\textbf{USER:} Can I be on Ele.me at the same time?\\

\textbf{Target 9}
\begin{lstlisting}[language=json, basicstyle=\ttfamily\small]
{
  "step": "8.2",
  "response": "Yes, you can. There are no restrictions on this.",
  "action": "moveon"
}
\end{lstlisting}

\textbf{Input 10}
\vspace{5px}

\textbf{ASSISTANT:} Hello, this is the Business Development Manager from Speedy Delivery Raccoon Canteen. May I ask if you are the owner of the store?\\
\textbf{USER:} This is Ma Jianqi, I am the owner of the store.\\
\textbf{ASSISTANT:} Hello, this is Speedy Delivery. We've just launched a new platform-supported food court with rent below market rates and priority traffic allocation - much more than regular takeout shops. We support both dine-in and delivery operations. Would you be interested in learning more?\\
\textbf{USER:} Okay.\\
\textbf{ASSISTANT:} Our main focus is on trustworthy food delivery. If you open a store with us, your shop will be labeled as a 'Trusted Store' on the Speedy Delivery app, which makes customers more willing to place orders. New stores will immediately receive an RMB 2,000 traffic bonus package, plus two weeks of traffic promotion (regular stores only receive one week). If your store performs well, the platform will provide additional traffic and list your dishes on other platform channels to help you earn more. Would you prefer a store with dine-in service or delivery-only?\\
\textbf{USER:} Either works.\\
\textbf{ASSISTANT:} Okay, I'll make a note of that. A dedicated business manager will follow up with you later. Is there anything else I can help you with for now\\
\textbf{USER:} Can I be on Ele.me at the same time?\\
\textbf{ASSISTANT:} Yes, you can. There are no restrictions on this.\\
\textbf{USER:} Can I make stir-fried dishes?\\

\textbf{Target 10}
\begin{lstlisting}[language=json, basicstyle=\ttfamily\small]
{
  "step": "8.3",
  "response": "Yes, you can. Our facilities are fully equipped to support a wide range of cooking needs.",
  "action": "moveon"
}
\end{lstlisting}

\end{tcolorbox}


\end{document}